\documentclass[preprint,12pt]{elsarticle}

\usepackage[colorlinks=true,linkcolor=blue,urlcolor=blue,citecolor=blue]{hyperref}
\usepackage{amssymb}
\usepackage{amsmath}
\usepackage{cleveref}
\usepackage{color}
\usepackage{url}
\usepackage{tcolorbox}
\usepackage{tabularx}
\usepackage{colortbl}
\usepackage{lineno}

\definecolor{green}{HTML}{44AA99}
\definecolor{yellow}{HTML}{DDCC77}
\definecolor{blue}{HTML}{88CCEE}
\definecolor{red}{HTML}{CC6677}
\definecolor{darkred}{HTML}{DC3220}

\usepackage{array}
\usepackage{listings}
\usepackage[T1]{fontenc}
\usepackage{multirow}
\usepackage{mwe}
\usepackage[export]{adjustbox}
\usepackage{tabularray}

\usepackage{enumitem}
\setlist[itemize]{noitemsep}

\journal{}

\begin{document}

\begin{frontmatter}

\title{Detecting Hope, Hate, and Emotion in Arabic Textual Speech and Multi-modal Memes Using Large Language Models}

\author[label1]{Nouar AlDahoul}
\author[label1]{Yasir Zaki\texorpdfstring{\corref{cor1}}{}}\ead{yasir.zaki@nyu.edu}

\cortext[cor1]{Corresponding author.}

\affiliation[label1]{organization=Computer Science Department, New York University Abu Dhabi,city=Abu Dhabi,country=UAE}

\begin{abstract}
 The rise of social media and online communication platforms has led to the spread of Arabic textual posts and memes as a key form of digital expression. While these contents can be humorous and informative, they are also increasingly being used to spread offensive language and hate speech. Consequently, there is a growing demand for precise analysis of content in Arabic text and meme. This paper explores the potential of large language models to effectively identify hope, hate speech, offensive language, and emotional expressions within such content.
We evaluate the performance of base LLMs, fine-tuned LLMs, and pre-trained embedding models . The evaluation is conducted using a dataset of Arabic textual speech and memes proposed in the ArabicNLP MAHED 2025 challenge. The results underscore the capacity of LLMs such as GPT-4o-mini, fine-tuned with Arabic textual speech, and Gemini Flash 2.5, fine-tuned with Arabic memes, to deliver the superior performance. They achieve up to 72.1\%, 57.8\%, and 79.6\% macro F1 scores for task 1, 2, and 3, respectively and secure first place overall in the challenge~\cite{Leaderboard_MAHED2025}. The proposed solutions offer a more nuanced understanding of both text and memes for accurate and efficient Arabic content moderation systems.
 
\end{abstract}


\end{frontmatter}

\section{Introduction}

AI content moderation refers to the use of artificial intelligence to monitor, evaluate, and manage content across digital platforms~\cite{verpex2023moderation}. By ensuring that posts comply with community standards and legal regulations, it helps create safer, more respectful, and law-abiding online environments. Its role has become increasingly vital as the volume and complexity of online content continue to grow. Despite growing efforts, Arabic content moderation still lags behind. Challenges such as dialect diversity, limited training data, and under-resourced tools make it difficult to ensure effective moderation across Arabic-speaking regions~\cite{campbell_singh_2021,tgi2023contentmoderation}.

Although Arabic is spoken by around 380 million people, it is far from being a uniform language~\cite{tgi2023contentmoderation}. It consists of six major regional dialect groups, so for classifiers to work effectively, they must be trained across all these dialects. The rise of social media and online communication platforms has led to the spread of Arabic
textual posts and memes as a key form of digital expression. There is a growing need to develop methods for detecting hateful text and memes, as they can perpetuate harmful stereotypes and contribute to the spread of offensive language and hate speech in digital spaces~\cite{zaghouani2024so,aldahoul2024advancing}. 

To have a full understanding of the emotional landscape of online communication, recognition of emotional expression can provide deeper insight into user sentiment and foster empathy. Additionally, emotional expression classification has valuable applications such as monitoring mental health and tailoring personalized recommendations~\cite{zaghouani2025emohopespeech}. 

Memes are especially widespread and can be potent tools for spreading propaganda, inciting hate, or conveying humor. LLMs have been shown to have superior performance in various domains and applications~\cite{aldahoul2025multitasking,aldahoul2024exploring}. For meme understanding, having textual and visual inputs, LLMs can analyze both the linguistic content and the underlying visual elements of a meme.

Our analyses and experiments center around the following research questions: \textbf{RQ1}: Can a pre-trained embedding model, combined with trained SVM or DNN classifiers, effectively detect hate and hope speech in Arabic text and memes? \textbf{RQ2}: Are existing safety classification and content moderation solutions capable of detecting hate speech in Arabic memes? \textbf{RQ3}: To what extent do state-of-the-art base LLMs excel in detecting hate speech in Arabic memes? \textbf{RQ4}: Can fine-tuned LLMs detect emotion, hope, hate, and offensive content in  Arabic text with high accuracy?
\textbf{RQ5}: Can fine-tuned LLMs detect hateful Arabic memes with high accuracy?

\section{Related Work}

Several studies have investigated hate speech and offensive language in Arabic text~\cite{mohaouchane2019detecting,kaddoura2023arabic,mubarak2023emojis,shapiro2022alexu,albadi2018they,bennessir2022icompass}. They utilized Convolutional Neural Networks (CNNs),  Long Short-Term Memory (LSTM), CNN-LSTM~\cite{mohaouchane2019detecting,kaddoura2023arabic}, word embeddings with simple Recurrent Neural Networks (RNN)~\cite{albadi2018they} and MARBERT~\cite{shapiro2022alexu,bennessir2022icompass}. The datasets used for analysis contain social posts and tweets.

To study the proportion of hate speech and offensive language in Arabic tweets, AraBERT was utilized~\cite{zaghouani2024so}. They found that 15\% of tweets contained offensive language, while 6\% included hate speech. Additionally, their annotated tweet dataset provided a valuable contribution to the limited availability of Arabic data related to hate speech and offensive language~\cite{zaghouani2024so}. It was found that AraBERT outperformed conventional machine learning classifiers ~\cite{zaghouani2024so}.

Even though there are several English emotion datasets, there is still a shortage of comprehensive Arabic datasets that support the analysis of both emotions and hope speech. \cite{zaghouani2025emohopespeech} proposed an Arabic dataset, fostering better cross-linguistic analysis of emotions and hope speech. They fine-tuned the AraBERT model~\cite{antoun2003arabert} for the hate-hope classification task. 

Building on previous research, numerous studies broadened the scope to tackle the challenge of detecting Arabic content across multiple modalities. In the context of Arabic propaganda identification~\cite{alam2024armeme,hasanain2024araieval}, separate feature extractors were employed for text and images. Moving from propaganda to hate, a multi-modal analysis of Arabic memes was done to further detect hate in memes. They used a fusion of features extracted from AraBERT for text and ConvNxT for images~\cite{alam2024propaganda}. 

\vspace{-5pt}
\section{Materials and Methods}
\vspace{-7pt}

\subsection{Dataset Overview}

Here we describe the datasets proposed in the ArabicNLP MAHED 2025 challenge~\cite{MAHED} that we utilized to run our experiments. 

\subsubsection{First dataset} 
This dataset is text-based speech that includes 9,843 examples for training, 1,476 for validation, and 1,477 for testing. The goal of using this data is to classify the speech text into one of three categories: hope, hate, or not\_applicable.  

\subsubsection{second dataset}
This dataset is a text-based multi-task set that contains 8,515 examples (5,960 for training, 1,277 for validation, and 1,278 for testing) and supports three types of sub-tasks.
The first sub-task aims to classify each text into one of twelve emotions: neutral, anger, anticipation, disgust, fear, joy, love, optimism, pessimism, sadness, surprise, or trust. The second sub-task aims to detect offensive language in the text, labeling it as either yes or no. When offensive language is detected, the third sub-task classifies the text as either hate or not\_hate.

\subsubsection{Third dataset}
\begin{figure}
    \centering
    \includegraphics[width=1\linewidth]{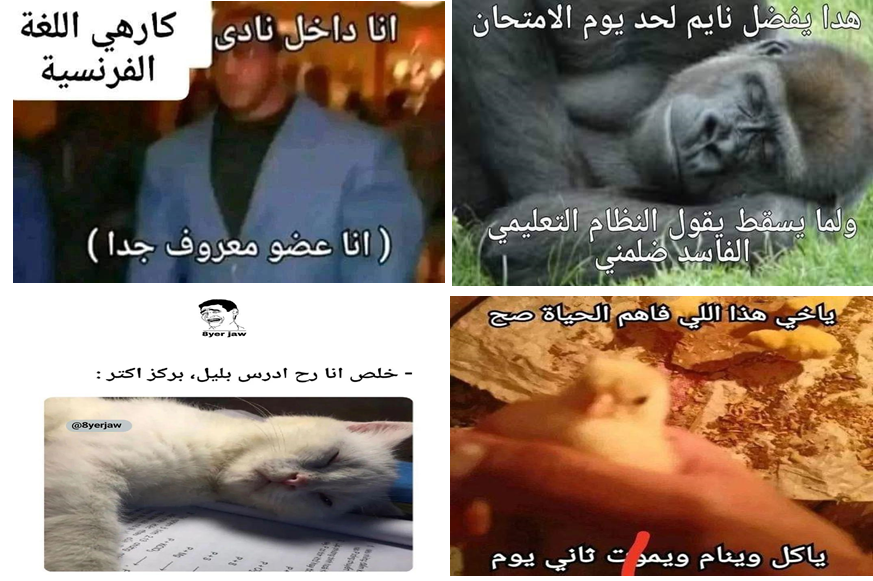}
    \caption{Examples from the third dataset.}
    \label{fig:data3}
\end{figure}

This dataset targets multi-modal hateful meme detection. It has 4,500 examples (2,143 for training, 312 for validation, and 606 for testing) annotated with two labels: hateful and not hateful. Each meme example includes an image and its extracted Arabic text. Examples from this dataset is shown in the Figure~\ref{fig:data3}.

\subsection{Methods}

\subsubsection{Detection of Hope and Hate in Arabic Speech}
For this task, first, we fine-tuned 2 LLMs such as  GPT-4o-mini~\cite{GPT-4o} (namely LLM 1 in Table~\ref{tab:task1}), and Gemini Flash 2.5~\cite{team2023gemini,gemini_25_thinking}(namely LLM 2 in Table~\ref{tab:task1}) using the training and validation sets from the \textbf{first dataset}. 
Secondly, we utilized Google   text embedding~\cite{google2025textmodal} + SVM~\cite{hearst1998support} (namely LLM 3 in Table~\ref{tab:task1}). To improve the accuracy, we used
ensemble learning (namely Ensemble in Table~\ref{tab:task1}) that used majority voting among previous 3 models. We found that many hope samples were predicted as not\_applicable. So we added hope/not\_applicable fine-tuned GPT-4o-mini to address this issue. we named our solution in Table~\ref{tab:task1}. 
 We reported the results of inference on a testing set.  The prompt used is as follows:

\begin{tcolorbox}[colback=orange!5!white,colframe=orange!75!black, title={Hate/Hope Detection Prompt}, rounded corners, boxrule=1pt, boxsep=1pt]
Classify each text into one of the three categories 'hope','hate',or 'not\_applicable'
\end{tcolorbox}

\subsubsection{Multi-task Detection of Emotional Expressions, Offensive Language, and Hate Speech}
For this task, three GPT-4o-mini models~\cite{GPT-4o} were fine-tuned using the training and validation sets from the \textbf{second dataset} for three epochs with a learning rate multiplier of 1.8. We reported the results of inference on a testing set. 

 To address the class imbalance in hate/not-hate sub-task, we over-sampled the minority `hate' text by a factor of five to achieve a more balanced distribution between the `hate' and `not-hate' classes.   The prompts used for each task are as follows:

\begin{tcolorbox}[colback=cyan!5!white,colframe=cyan!75!black, title={Offensive Language Detection Prompt}, rounded corners, boxrule=1pt, boxsep=1pt]
You are an expert in offensive language detection.\\
Carefully read the input text and determine whether it is offensive.\\
Respond with only one word, either "yes" if the text is 
offensive, or "no" if it is not.\\
Text: "<your text here>"\\
Answer:
\end{tcolorbox}

\begin{tcolorbox}[colback=cyan!5!white,colframe=cyan!75!black, title={Hate/not-Hate Detection Prompt}, rounded corners, boxrule=1pt, boxsep=1pt]
Classify each text into one of the two categories 'hate', or 'not\_hate'
\end{tcolorbox}

\begin{tcolorbox}[colback=cyan!5!white,colframe=cyan!75!black, title={Emotion Detection Prompt}, rounded corners, boxrule=1pt, boxsep=1pt]
You are an expert in fine-grained emotion classification.\\
Given a short piece of text, your task is to classify the **dominant emotion** expressed in the text.\\
Use **only one word** from the following list of predefined emotion categories:\\

[anger, disgust, neutral, love, joy, anticipation, optimism, sadness, confidence, pessimism, surprise, fear]\\
Carefully read the input and select the most appropriate emotion based on the **overall tone, context, and word choice**.\\
If multiple emotions are expressed, choose the one that is **most central** to the message.\\
Respond with only one word, exactly as listed.\\
Text: "<your text here>"\\
Answer:\\
\end{tcolorbox}

\subsubsection{Multi-modal Detection of Arabic Hateful Memes}
For this task, we have evaluated several methods, including base LLMs, fine-tuned LLMs, and embedding models, to find the best solution. We tested all solutions using the testing data of 606 Arabic memes.  The prompts used are as follows:

\begin{tcolorbox}[colback=purple!5!white,colframe=purple!75!black, title={Hate Detection in Memes Prompt 1}, rounded corners, boxrule=1pt, boxsep=1pt]
You're an expert Arabic social media hate content analyst.\ Your goal is to flag any content that is derogatory, disparaging, promotes negative stereotypes, incites any form of harm (even humorously), or contributes to a hostile environment.\\
Provide the following for analysis:\\
Meme Description (Visuals): Detailed text description of all visual elements (characters, symbols, actions, etc.).\\
Meme Text (Arabic \& English): Exact Arabic text and its literal English translation. Include any emojis/symbols.\\
Context: Explain who is targeted (individuals, groups, professions, etc.), relevant current events,
cultural/historical references, and any known slang/derogatory terms.\\
User Intent (if known): Any information on the creator/sharer's likely purpose.\\
Analysis: {Your detailed reasoning, explaining how visuals, text, and context contribute to hate.\\
Focus on dehumanization, negative stereotypes, incitement (even if satirical), hostile environment creation, contempt, ridicule, disgust, targeting based on role, disparaging language (ableism, body shaming, etc.),
and normalization of problematic behavior.}\\
Final Answer: {hate/no hate}
\end{tcolorbox}

\begin{tcolorbox}[colback=blue!5!white, colframe=blue!75!black, title={Hate Detection in Memes Prompt 2}, rounded corners, boxrule=1pt, boxsep=1pt]
You are a hate speech analyst evaluating Arabic social media memes.\\
Your goal is to flag hate speech, including implicit hate masked as humor or political discourse.\\
Hate includes:\\
Stereotyping or mocking a group based on ethnicity, nationality, gender, religion, or identity
Jokes or memes that promote harmful generalizations\\
Political content that blames or dehumanizes a people, not just a government or policy\\
However, you must not flag:\\
Political critique focused on leadership, government, or policy (not people)\\
Cultural observations or satire without negative generalization \\
Your analysis must:\\
Describe the image and interpret the Arabic text\\
Explain whether it includes group-based bias or stereotypes\\
End with:\\
Analysis: {your reasoning}\\
Final Answer: {'hate' or 'no hate'}\\
\end{tcolorbox}

\begin{tcolorbox}[colback=green!5!white, colframe=green!75!black, title={Hate Detection in Memes Prompt 3}, rounded corners, boxrule=1pt, boxsep=1pt]
You are a hate speech analyst evaluating Arabic social media memes.\\
Your goal is to classify meme into hate or no\_hate
\end{tcolorbox}

First, we started with assessing the performance of \textbf{embedding models} that can combine their outputs with traditional classifiers for hate/not-hate classification. We used the Google multi-modal pre-trained embedding model (multimodalembedding@001)~\cite{google2025multimodal} to generate embedding vectors for each text and image in each meme. The embedding vector has 512 dimensions. Later, we aggregated the two embedding vectors of text and image by computing their element-wise average first and then by concatenating the two vectors. Finally, we added a support vector machine (SVM)~\cite{hearst1998support} to classify the resulting embedding vector into two classes: hate and not-hate. We assessed four scenarios: text embedding vector only, image embedding vector only, average of text and image embedding vectors, and concatenation of text and image embeddings. We fine-tuned hyperparameters of SVM to get the highest F1 and F2 scores. We found that regularization parameter C = 0.1, kernel = radial basis function (rbf), gamma = scale, and balanced class weighted loss function are the optimal hyperparameters for the three scenarios except the text-only scenario, where C = 1 is optimal. Additionally, we replaced SVM with a deep neural network (DNN)~\cite{lecun2015deep} whose architecture was optimized to get the optimal one with the highest F1 and F2 scores. The architecture is shown in Table~\ref{tab:dnn-architecture}. The hyper-parameters of DNN are as follows:

\begin{tcolorbox}[colback=gray!5!white, colframe=gray!75!black, title={Hyper-parameters for DNN}, rounded corners, boxrule=1pt, boxsep=1pt]
\begin{itemize}
\item Adam optimizer,
\item weighted class binary cross-entropy loss fuction
\item 100 epochs
\item 128 batch size
\item early stopping with patience = 3. 
\end{itemize}
\end{tcolorbox}

\begin{table}[!ht]
\centering
\caption{\normalsize Architecture of the dual-branch DNN model for image and text fusion.}
\resizebox{\columnwidth}{!}{%
\begin{tabular}{cccc}
\hline
\textbf{Layer Type} & \textbf{Output Shape} & \textbf{Activation} & \textbf{Description} \\ \hline
Input Layer         & (512,)                & –                   & Input vector representing image embedding  \\ 
Dense Layer         & (256,)                & ReLU                & Fully connected layer on image input       \\ 
Dropout             & (256,)                & Dropout 0.5         & Regularization                             \\ 
Dense Layer         & (128,)                & ReLU                & Further transformation of image embedding  \\ 
Dropout             & (128,)                & Dropout 0.5         & Regularization                             \\ 
Dense Layer         & (64,)                 & ReLU                & Compressed feature representation          \\ 
Dropout             & (64,)                 & Dropout 0.5         & Regularization                             \\ 
Input Layer         & (512,)                & –                   & Input vector representing text embedding   \\ 
Dense Layer         & (256,)                & ReLU                & Fully connected layer on text input        \\ 
Dropout             & (256,)                & Dropout 0.5         & Regularization                             \\ 
Dense Layer         & (128,)                & ReLU                & Intermediate transformation                \\ 
Dropout             & (128,)                & Dropout 0.5         & Regularization                             \\ 
Dense Layer         & (64,)                 & ReLU                & Compressed feature representation          \\ 
Dropout             & (64,)                 & Dropout 0.5         & Regularization                             \\ 
Concatenate         & (128,)                & –                   & Merge image and text features (64 + 64)    \\ 
Dense Layer         & (128,)                & ReLU                & Combined representation processing         \\ 
Dropout             & (128,)                & Dropout 0.5         & Regularization                             \\ 
Dense Layer         & (1024,)               & ReLU                & High-capacity layer for rich interaction   \\ 
Dropout             & (1024,)               & Dropout 0.5         & Regularization   \\ 
Dense Layer         & (1,)                  & Sigmoid             & Final prediction for binary classification \\ \hline
\end{tabular}%
}
\label{tab:dnn-architecture}
\end{table}

In the \textbf{second experiment}, we assessed the capacity of multi-modal pre-trained \textbf{safety classifiers} for hate detection in memes. We started with \textbf{Llama Guard 4}~\cite{Llama_Guard_vision,llama_guard_4} which is a multi-modal safety classifier with 12 billion parameters, trained jointly on both text and images. It uses a dense architecture derived from the Llama 4 Scout pre-trained model, which has been pruned and fine-tuned specifically for content safety classification. In this work, our focus is on the `hate' category, which refers to text that demeans or dehumanizes individuals based on sensitive personal characteristics. We focus on all examples that have been flagged under the hate category only.

Then we assessed \textbf{Omni-moderation-latest~\cite{OpenAI_Moderation}} which is a moderation endpoint used to check whether text or images are potentially harmful. Its output includes several categories and their confidence values. The moderator sets the flag to true if it classifies the content as harmful. The limitation of this is that for categories such as `hate' or `hate/threatening,' it supports only text. We consider all examples that have triggered the safety flag.

In the \textbf{third experiment}, we ran \textbf{Gemini Flash 2.5}~\cite{team2023gemini,gemini_25_thinking}, a base model with a system prompt (Prompt 1). We also ran the \textbf{GPT-4o-mini} base model with Prompts 1, 2, and 3.
 
To improve the detection performance, we fine-tuned several LLMs in a supervised learning setting. We started by tuning Gemini Flash 2.5 using Prompt 3. To address the class imbalance, we over-sampled the minority `hate' memes by a factor of nine to achieve a more balanced distribution between the `hate' and `no\_hate' classes. The hyper-parameters used are three epochs, learning\_rate\_multiplier of 0.5, an adapter size of 2, an off threshold in safety\_settings, and disabled thinking. Additionally, we also fine-tuned \textbf{Llama 3.2-11B~\cite{Llama_vision}} using both text and image inputs from the training data. We used Low-Rank Adaptation (LoRA)~\cite{hu2022lora} as the Parameter-Efficient Fine-Tuning (PEFT)~\cite{xu2023parameter} method for fine-tuning utilizing the unsloth framework.
The fine-tuned Llama 3.2-11B model was uploaded to Huggingface:
\url{https://huggingface.co/NYUAD-ComNets/Llama3.2-MultiModal-Hate\_Detector\_Memes}
The hyper-parameters of Llama 3.2-11B are as follows:

\begin{tcolorbox}[colback=gray!5!white, colframe=gray!75!black, title={Fine-tuning Configurations of Llama 3.2-11B}, rounded corners, boxrule=1pt, boxsep=1pt]
\begin{itemize}
\item the training batch size per device is set to 4.
\item gradients are accumulated over 4 steps.
\item the learning rate warm-up lasts for 5 steps.
\item the total number of training steps is 150.
\item the learning rate is set to 0.0002.
\item the optimizer used is 8-bit AdamW
\item weight decay is set to 0.01.
\item a linear learning rate scheduler is used.
 \end{itemize}
\end{tcolorbox}

Finally, we fine-tuned \textbf{Paligemma2~\cite{paligemma_3b_pt_224,steiner2412paligemma}} 
namely ``google/paligemma2-3b-pt-224''. The parameters of the vision tower and the language model are frozen, while only the parameters of the multi-modal projector are set to be trainable. 
The hyper-parameters of Paligemma2 are as follows:

\begin{tcolorbox}[colback=gray!5!white, colframe=gray!75!black, title={Training Configuration of PaliGemma2}, rounded corners, boxrule=1pt, boxsep=1pt]
\begin{itemize}
\item number of training epochs: 3
\item per-device training batch size: 2
\item gradient accumulation steps: 8
\item warm-up steps: 2
\item learning rate: 2e-5
\item weight decay: 1e-6
\item Adam optimizer beta2 value: 0.999
\item optimizer type: Adamw\_hf
\item early stopping callback with patience=2.
\end{itemize}
\end{tcolorbox}

In the previous fine-tuning experiments, we used OpenAI~\cite{GPT-4o} for tuning each GPT-4o-mini.  Additionally, we used Google AI vertex platform~\cite{Gemini_Vertex} for tuning Gemini Flash 2.5.

\section{Results and Discussion}

\subsection{Hate/Hope Detection in textual speech}
\begin{figure}[!htb]
    \centering
    \includegraphics[width=0.8\linewidth]{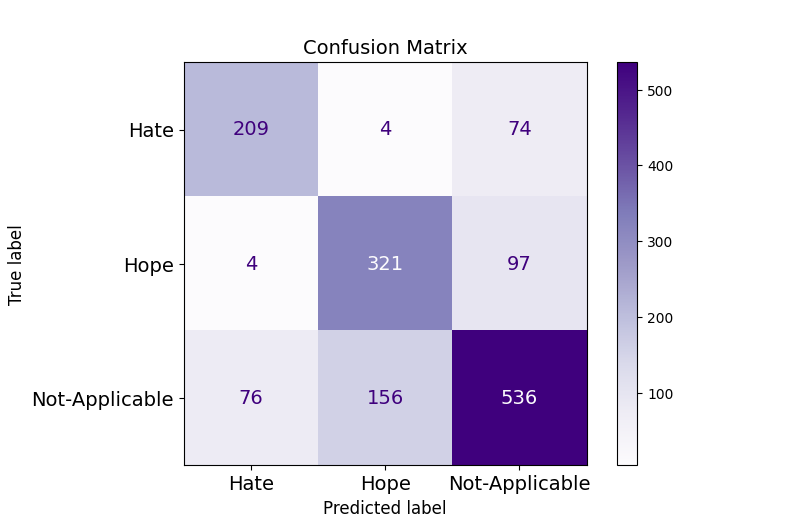}
    \caption{Confusion matrix of testing set in task 1 for hope/hate/not\_applicable classification in text using ensemble of 3 fine-tuned LLMs (gpt-4o-mini, Gemini Flash 2.5, and Google text embedding + SVM) + fine-tuned gpt-4o-mini for hope/not.}
    \label{fig:conf_mat_hop-hate}
\end{figure}

In this task,  the ensemble method of majority voting among the three LLMs improved the performance as shown in Table~\ref{tab:task1}. Moreover, adding the hope/not classifier to better distinguish real hope samples from those predicted as not\_applicable achieved the best performance metrics and ranked second in the leaderboard~\cite{Leaderboard_MAHED2025} which addresses \textbf{RQ4}. It is also worth mentioning that embedding model + SVM (LLM3) shows good performance which answers \textbf{RQ1}. Figure~\ref{fig:conf_mat_hop-hate} depicts the confusion matrix of testing set in task 1 for hope/hate/not\_applicable classification in text using ensemble of 3 fine-tuned LLMs (gpt-4o-mini, Gemini Flash 2.5, and Google text embedding + SVM) + fine-tuned gpt-4o-mini for hope/not.

\begin{table}[!ht]
\centering
\caption{\normalsize Performance metrics for Task 1 (hop/hate/not\_applicable).}
\footnotesize
\begin{tabularx}{\columnwidth}{m{1.35cm}XXXX}
\hline
\textbf{Task} & \textbf{Accuracy \%} & \textbf{Macro Precision \%} & \textbf{Macro Recall \%} & \textbf{Macro F1 Score \%} \\ \hline

\textbf{LLM 1}         & 70.6           & 70.6                     & 69                    & 69.7                    \\ \hline
\textbf{LLM 2}    & 69.7            & 68.6                     & 72.2                 & 69.9                   \\ \hline
\textbf{LLM 3}    & 70.6           & 71.6                    & 67.2                  & 68.9                    \\ \hline
\textbf{Ensemble}                         & 71.9          & 71.7                     & 71.2                  & 71.4                   \\ \hline
\textbf{Our Solution} & \textbf{72.3}            & 71.6                     & \textbf{72.9}                  & \textbf{72.1}                    \\ \hline

\end{tabularx}%

\label{tab:task1}
\end{table}

\begin{figure}[!htb]
    \centering
    \includegraphics[width=0.8\linewidth]{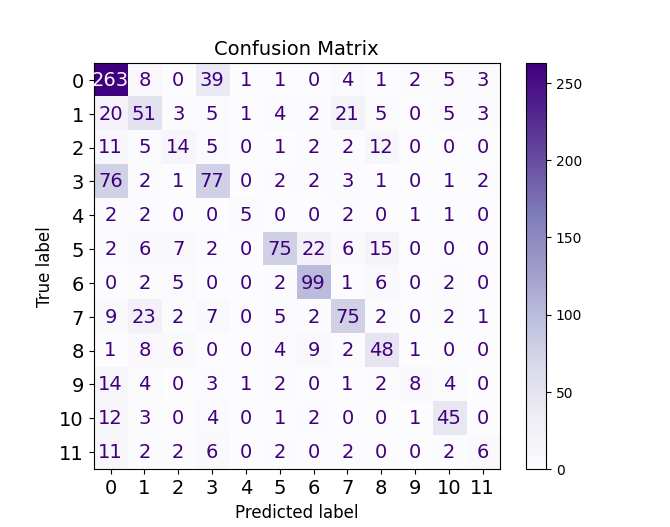}
    \caption{Confusion matrix of testing set in task 2 for emotion classification in text using Fine-tuned GPT-4o-mini. class 0: Anger, class 1: Anticipation, class 2: Confidence, class 3: Disgust, class 4: Fear, class 5: Joy, class 6: Love, class 7: Neutral, class 8: Optimism, class 9: Pessimism, class 10: Sadness , class 11: Surprise.}
    \label{fig:conf_mat_emotion}
\end{figure}

\subsection{Multi-task Detection}
\begin{figure}[!htb]
    \centering
    \includegraphics[width=0.8\linewidth]{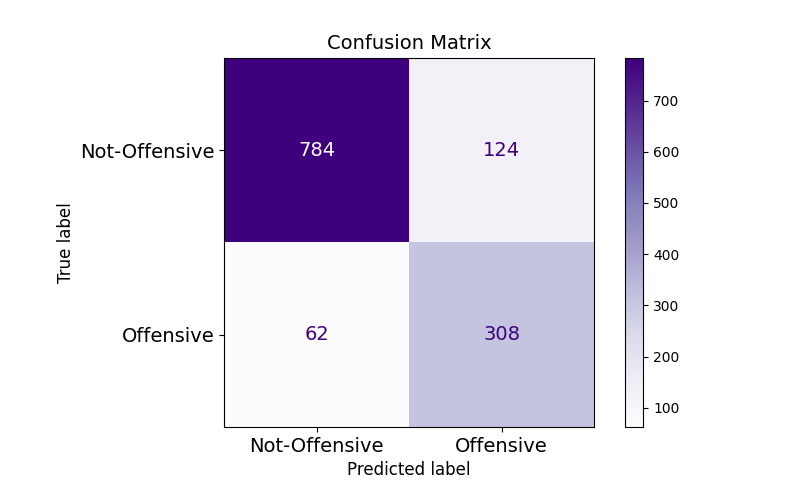}
    \caption{Confusion matrix of testing set in task 2 for offensive detection in text using Fine-tuned GPT-4o-mini.}
    \label{fig:conf_mat_offensive}
\end{figure}

The three fine-tuned GPT-4o-mini for multi-task (emotion, offensive, hate) achieved the best performance compared to other methods in the leaderboard~\cite{Leaderboard_MAHED2025} evaluated on a testing set which addresses \textbf{RQ4}. More details in Table~\ref{tab:task2}. The model achieved a Macro F1-score of 57.8\%, an accuracy of 75.0\%, a precision of 61.2\%, and a recall of 57.8\% over all three sub-tasks. Figure~\ref{fig:conf_mat_emotion} shows confusion matrix of testing set in task 2 for emotion classification in text using Fine-tuned GPT-4o-mini. Additionally, Figure~\ref{fig:conf_mat_offensive} illustrates confusion matrix of testing set in task 2 for offensive detection in text using Fine-tuned GPT-4o-mini. 

\begin{table}[ht]
\centering
\caption{performance metrics for multi-task (task 2). Hate/Not detection is influenced by the offensive detection step, and some evaluation metrics cannot be computed because samples predicted as non-offensive yield NaN values and are excluded from the Hate/Not detector.}
\footnotesize
\begin{tabularx}{\columnwidth}{m{1.4cm}XXXXX}
\hline
\textbf{Task}          & \textbf{Accuracy \%} & \textbf{Macro Precision \%} & \textbf{Macro   Recall \%} & \textbf{Macro F1 score \%} \\ \hline
\textbf{Emotion}       & 59.9            & 57.2                     & 49.9                  & 51.7                   \\ \hline
\textbf{Offensive/ Not} & 85.4            & 82.0                     & 84.8                  & 83.1                    \\ \hline
\textbf{Hate/Not}      & 63.8            & -                          & -                       & -                         \\ \hline
\end{tabularx}

\label{tab:task2}
\end{table}

\subsection{Hate Detection in Memes} Table~\ref{tab:llm-comparison}  presents performance metrics for a variety of models. A pre-trained multi-modal embedding model was found to effectively detect hate speech in Arabic memes using either SVM or DNN, answering \textbf{RQ1}. Both LLaMA 4 Guard and OpenAI content moderator show lower recall and F1-scores, especially OpenAI one, suggesting limitations in the existing safety classification solutions on this task, which  addresses \textbf{RQ2}. Among the base LLMs, GPT-4o demonstrated stronger performance compared to Gemini Flash 2.5, answering \textbf{RQ3}.

\begin{figure}[!htb]
    \centering   \includegraphics[width=0.7\linewidth]{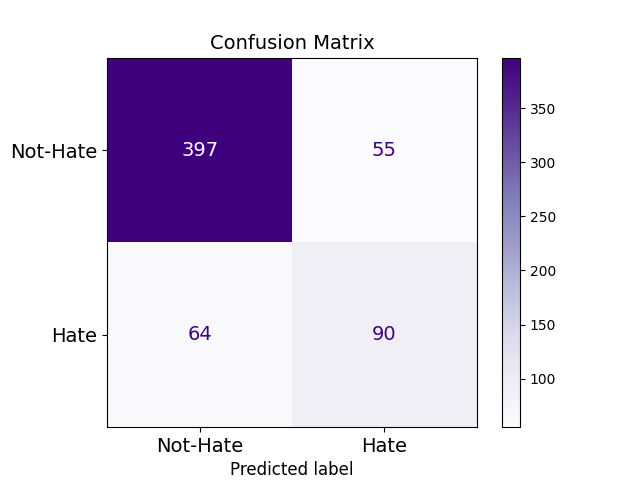}
    \caption{Confusion matrix of validation set for hate detection in memes using Fine-tuned Llama 3.2 11B.}
    \label{fig:conf_mat_p7}
\end{figure}

Fine-tuned Gemini Flash 2.5 demonstrates superior performance across all metrics. Similarly, fine-tuned Llama 3.2 11B consistently ranks second. Figure~\ref{fig:conf_mat_p7} shows confusion matrix of testing set for hate detection in memes using Fine-tuned
Llama 3.2 11B.  The results indicate that fine-tuning significantly boosts models' capabilities, which addresses \textbf{RQ5}. On the other hand, fine-tuned PaliGemma2 underperforms compared to other models. 

\begin{figure}[!htb]
    \centering   \includegraphics[width=0.7\linewidth]{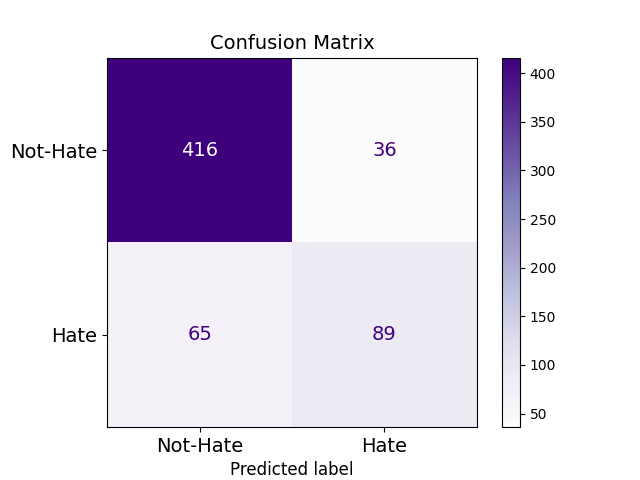}
    \caption{Confusion matrix of validation set for hate detection in memes using Fine-tuned Gemini Flash 2.5.}
    \label{fig:conf_mat_p6}
\end{figure}

\begin{figure}[!htb]
    \centering
    \includegraphics[width=0.7\linewidth]{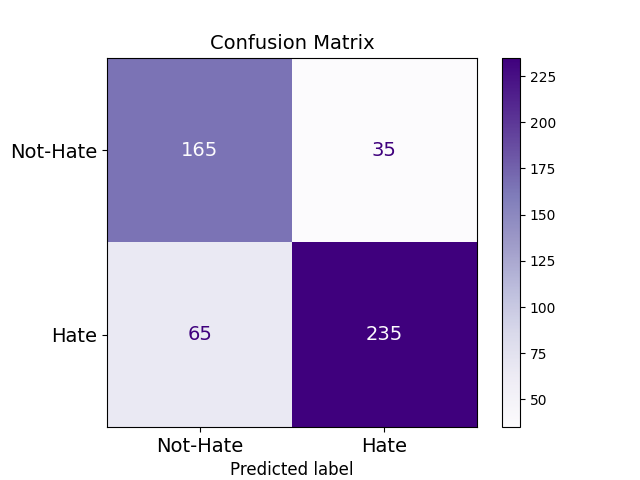}
    \caption{Confusion matrix of testing set for hate detection in memes using Fine-tuned Gemini Flash 2.5.}
    \label{fig:conf_mat_test}
\end{figure}

Flash Flash 2.5 achieved the best performance in the leaderboard~\cite{Leaderboard_MAHED2025} evaluated on a testing set of 500 memes. The model achieved a Macro F1-score of 79.6\%, an accuracy of 80.0\%, a precision of 79.4\%, and a recall of 80.4\%. 
Figure~\ref{fig:conf_mat_p6}, and Figure~\ref{fig:conf_mat_test} show confusion matrix of validation set and testing set for hate detection in memes using Fine-tuned
Gemini Flash 2.5. 

\begin{table}[ht]
\caption{Performance of base and fine-tuned LLMs for task 3.}
\centering
\footnotesize
\begin{tabularx}{\columnwidth}{m{1.5cm}XXXXX}
\hline
\textbf{LLM} & \textbf{Accuracy \%} & \textbf{Macro Precision \%} & \textbf{Macro Recall \%} & \textbf{Macro F1 score \%} & \textbf{Macro F2 score \%} \\ \hline
embedding + SVM & 77.56 & 70.48 & 70.83 & 70.65 & 70.76 \\ \hline
embedding + DNN & 77.56 & 70.32 & 69.97 & 70.14 & 70.04 \\ \hline
OpenAI content moderator & 72.77 & 57.27 & 52.85 & 51.18 & 51.75 \\ \hline
Llama 4 Guard & 71.45 & 63.36 & 64.38 & 63.77 & 64.11 \\ \hline
Base GPT-4o-mini & 79.21 & 72.49 & 71.29 & 71.84 & 71.50 \\ \hline
Base Gemini Flash 2.5 & 64.19 & 62.47 & 66.36 & 61.09 & 63.20 \\ \hline
Fine-tuned Gemini Flash 2.5 & \textbf{83.33} & \textbf{78.84} & \textbf{74.91} & \textbf{76.49} & \textbf{75.46} \\ \hline
Fine-tuned Llama 3.2 11B & 80.36 & 74.09 & 73.14 & 73.58 & 73.31 \\ \hline
Fine-tuned Paligemma2 & 76.73 & 68.95 & 67.49 & 68.12 & 67.72 \\ \hline
\end{tabularx}%

\label{tab:llm-comparison}
\end{table}

\newpage

\begin{figure}[!htb]
    \centering   \includegraphics[width=0.7\linewidth]{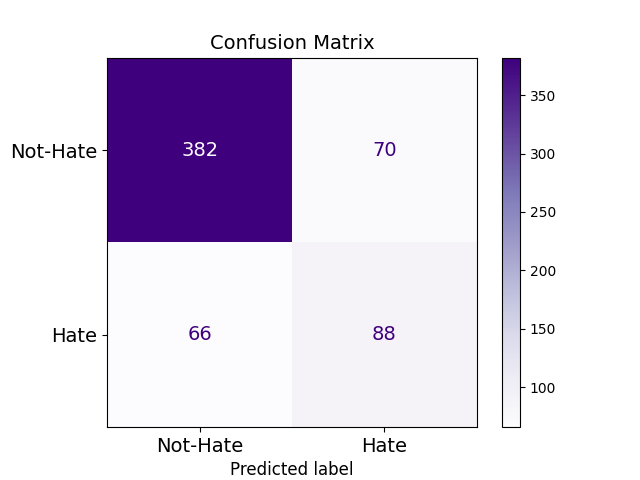}
    \caption{Confusion matrix of validation set for hate detection in memes using average embeddings of image and text + SVM.}
    \label{fig:conf_mat_p4}
\end{figure}

\begin{figure}[!htb]
    \centering   \includegraphics[width=0.7\linewidth]{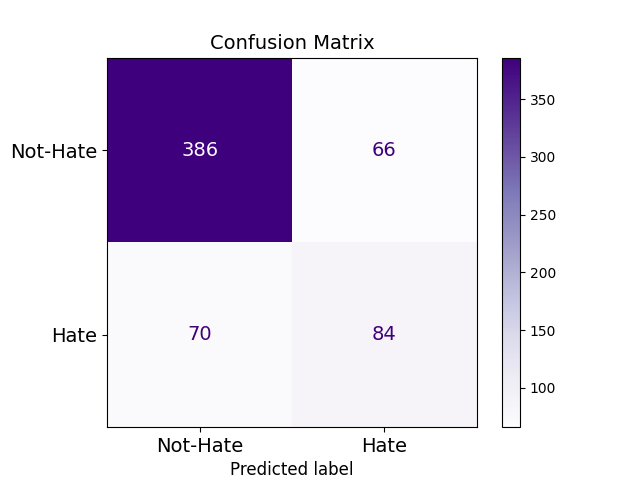}
    \caption{Confusion matrix of validation set  for hate detection in memes using average embeddings of image and text + DNN.}
    \label{fig:conf_mat_p5}
\end{figure}

Table~\ref{tab:svm-modalities} shows Google's multi-modal embedding model results with SVM for different input modalities. The findings indicate that the average embedding vector outperforms slightly the image-only embedding. This suggests that adding text embeddings does not provide an advantage for classification. One explanation is that Google's embedding model processes the text within the meme's image. The performance of text-only embeddings is the lowest. Figure~\ref{fig:conf_mat_p4}, and Figure~\ref{fig:conf_mat_p5} show confusion matrix of Google’s multi-modal embedding model results with SVM and DNN for hate detection in memes. 

\newpage
\begin{table}[!ht]
\centering
\caption{Performance of different input modalities combined with evaluated on validation set in task 3.}
\footnotesize
\begin{tabularx}{\columnwidth}{m{1.4cm}XXXXX}
\hline
 & \textbf{Accuracy \%} & \textbf{Macro Precision \%} & \textbf{Macro Recall \%} & \textbf{Macro F1 score \%} & \textbf{Macro F2 score \%} \\ \hline
image embedding + SVM         & 77.23 & 70.10 & 70.61 & 70.34 & 70.50 \\ \hline
text embedding + SVM         & 66.50 & 57.15 & 57.64 & 57.33 & 57.50 \\ \hline
Avg. embedding of image\&text + SVM & \textbf{77.56} & \textbf{70.48} & \textbf{70.83} & \textbf{70.65} & 
\textbf{70.76} \\ \hline
Concatenate embedding of image\&text + SVM  & 76.07 & 68.53 & 68.76 & 68.64 & 68.71 \\ \hline
\end{tabularx}%

\label{tab:svm-modalities}
\end{table}

\begin{figure}[!htb]
    \centering
    \includegraphics[width=0.7\linewidth]{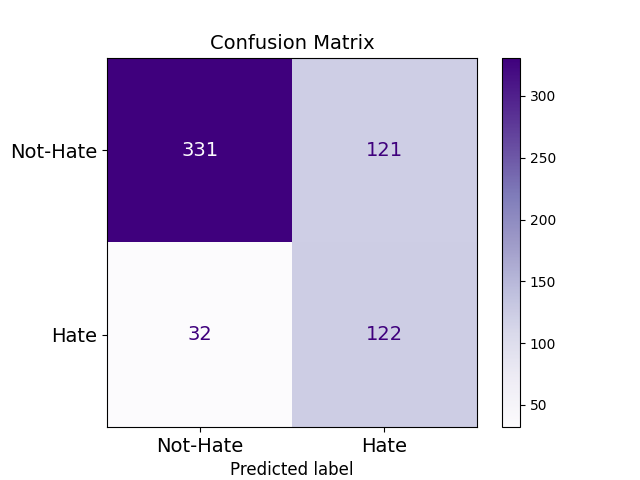}
    \caption{Confusion matrix of validation set for hate detection in memes using GPT-4o-mini with Prompt 1.}
    \label{fig:conf_mat_p1}
\end{figure}

\begin{figure}[!htb]
    \centering    \includegraphics[width=0.7\linewidth]{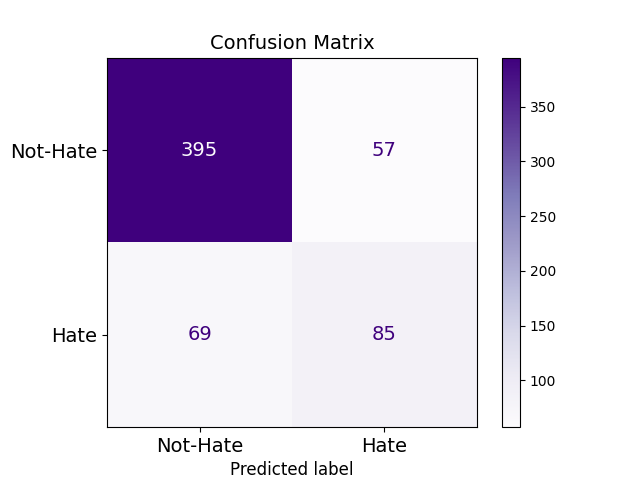}
    \caption{Confusion matrix of validation set for hate detection in memes using GPT-4o-mini with Prompt 2.}
    \label{fig:conf_mat_p2}
\end{figure}

\begin{figure}[!htb]
    \centering   \includegraphics[width=0.7\linewidth]{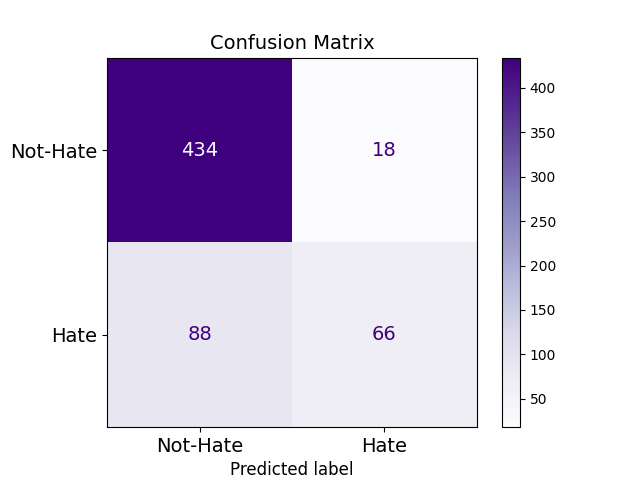}
    \caption{Confusion matrix of validation set for hate detection in memes using GPT-4o-mini with Prompt 3.}
    \label{fig:conf_mat_p3}
\end{figure}

We also ran base GPT-4o-mini with the three prompts as shown in Table~\ref{tab:GPT-4o-mini}. Even though Prompt 3 produced the highest accuracy and macro F1 score, Prompt 1 gave the highest macro F2 score, suggesting a better prompt to detect the hate class specifically.
Figure~\ref{fig:conf_mat_p1}, Figure~\ref{fig:conf_mat_p2}, and Figure~\ref{fig:conf_mat_p3} show confusion matrix of base gpt-4o-mini with each prompt  for hate detection in memes. 

\newpage
\begin{table}[!ht]
\centering
\caption{GPT-4o-mini base model performance under different prompts evaluated on validation set in task 3.}
\footnotesize
\begin{tabularx}{\columnwidth}{m{1.7cm}XXXXX}
\hline
 & \textbf{Accuracy \%} & \textbf{Macro Precision \%} & \textbf{Macro Recall \%} & \textbf{Macro F1 score \%} & \textbf{Macro F2 score \%} \\ \hline
GPT-4o-mini Prompt 1  & 74.75 & 70.70 & \textbf{76.23} & 71.34 & \textbf{73.62} \\ \hline
GPT-4o-mini Prompt 2  & 79.21 & 72.49 & 71.29 & 71.84 & 71.50 \\ \hline
GPT-4o-mini Prompt 3 & \textbf{82.51} & \textbf{80.86} & 69.44 & \textbf{72.29} & 70.14 \\ \hline
\end{tabularx}%

\label{tab:GPT-4o-mini}
\end{table}

\newpage

\newpage

\section*{Limitations}

One limitation of this work is the subjective nature of the annotations poses challenges, as different annotators may interpret and label content differently. This introduces potential inconsistencies in the training data, which could affect the model's performance.

Another key limitation is the models' ability to understand and process different Arabic dialects.

\bibliography{anthology,custom}
\bibliographystyle{acl_natbib}

\end{document}